\documentclass[journal]{IEEEtran}

\usepackage{soul}
\usepackage{subfigure}
\usepackage{amsmath}
\usepackage{subfigure}
\usepackage{graphicx}
\usepackage{color}
\usepackage{dblfloatfix}
\usepackage[framemethod=tikz]{mdframed}

\newcommand{\fig}[1]{Fig.~\ref{fig:#1}}
\newcommand{\eq}[1]{(\ref{eq:#1})}

\ifCLASSINFOpdf
\else
\fi

\begin{document}
\markboth{Journal of \LaTeX\ Class Files,~Vol.~14, No.~8, August~2015}%
{Shell \MakeLowercase{\textit{et al.}}: Bare Demo of IEEEtran.cls for IEEE Journals}
\title{SNR optimization of multi-span fiber optic communication systems employing EDFAs with non-flat gain and noise figure}

\author{Metodi Plamenov Yankov,~\IEEEmembership{Member,~IEEE}, Pawel Marcin Kaminski, Henrik~Enggaard~Hansen, Francesco Da Ros~\IEEEmembership{Member,~OSA, Senior Member,~IEEE,}%

\thanks{The authors are with the Department of Photonics Engineering, Technical University of Denmark, 2800 Kgs. Lyngby, Denmark, meya@fotonik.dtu.dk}}

\maketitle

\begin{abstract}
Throughput optimization of optical communication systems is a key challenge for current optical networks. The use of gain-flattening filters (GFFs) simplifies the problem at the cost of insertion loss, higher power consumption and potentially poorer performance. In this work, we propose a component wise model of a multi-span transmission system for signal-to-noise (SNR) optimization. A machine-learning based model is trained for the gain and noise figure spectral profile of a C-band amplifier without a GFF. The model is combined with the Gaussian noise model for nonlinearities in optical fibers including stimulated Raman scattering and the implementation penalty spectral profile measured in back-to-back in order to predict the SNR in each channel of a multi-span wavelength division multiplexed system. All basic components in the system model are differentiable and allow for the gradient descent-based optimization of a system of arbitrary configuration in terms of number of spans and length per span. When the input power profile is optimized for flat and maximized received SNR per channel, the minimum performance in an arbitrary 3-span experimental system is improved by up to 8 dB w.r.t. a system with flat input power profile. An SNR flatness down to 1.2 dB is simultaneously achieved. The model and optimization methods are used to optimize the performance of an example core network, and 0.2 dB of gain is shown w.r.t. solutions that do not take into account nonlinearities. The method is also shown to be beneficial for systems with ideal gain flattening, achieving up to 0.3 dB of gain w.r.t. a flat input power profile.
\end{abstract}


\section{Introduction}

Optical fiber links periodically amplified by Erbium doped fiber amplifiers (EDFAs) are one of the fundamental building blocks of telecommunication networks and are thus a key enabler of our digital society. The full C-band window of EDFAs allows for simultaneous amplification of all wavelength division multiplexed (WDM) channels propagating through an optical fiber. Standard EDFAs exhibit a non-flat gain spectral profile and gain competition between channels, resulting in input power spectral density (PSD)-dependent gain profile and eventually non-uniform amplification per user. Furthermore, the noise performance of EDFAs is also non-flat, due to non-spectrally flat noise figure (NF). These effects, combined with the PSD-dependent fiber nonlinearities, such as the Kerr effect and the stimulated Raman scattering (SRS) effect, result in a WDM performance which is non-uniform per user/channel and difficult to predict \cite{Zirngibl}. 

Modeling and optimization of fiber nonlinearities is typically approached with a generalized Gaussian noise (GGN) model \cite{Roberts}\cite{Semrau}. However, these works do not take into account the non-flatness of the EDFA gain and NF profiles. Flat gain can potentially be achieved by periodically applying extra amplification and a gain flattening filter (GFF) \cite{TorinoECOC2020}\cite{Perin}. This comes at the cost of energy consumption and further enhances the non-flatness of the accumulated noise profile, and potentially degrades the overall performance. More recently a few studies have started considering the EDFA response \cite{JuhnoECOC2020}\cite{JuhnoChoJLT}\cite{Maria}, but limit the analysis to systems operating in the linear region of the fiber. This is reasonable for transoceanic systems with electrical power constraint, but is sub-optimal for regional and terrestrial networks where maximizing the throughput by operating at the optimum launch power is essential. Such links are also typically parts of mesh networks, for which equalizing the performance per channel is preferred to maximizing the total throughput. Furthermore, the throughput is targeted for optimization only by proxy of the optical SNR (OSNR) in \cite{JuhnoECOC2020}\cite{JuhnoChoJLT}\cite{Maria}, which neglects potential frequency dependent transmitter and receiver penalties, as well as the nonlinear distortions accumulated during transmission. 

This paper is a direct extension of \cite{YankovJLT}, where the EDFA modeling and the system optimization were restricted to the power evolution. Here, the noise contributions from the EDFAs, fiber nonlinearities (Kerr and Raman scattering) and transmitter and receiver penalties are included at each corresponding component to construct an accurate cascaded model for the SNR of a \textit{completely reconfigurable} multi-span system. The power optimizations based on this model are experimentally validated on arbitrarily chosen link configurations in terms of fiber lengths and EDFA orders. To our knowledge, this is the first experimental demonstration of gain flattening-free transmission that accounts for the 1) EDFA non-flat and input-dependent NF and gain profile; 2) Kerr and SRS nonlinearities in standard, single mode fibers (SSMFs); and 3) transmitter and receiver implementation penalties. The optimization method is finally validated in a simulation of an example core network, where an analysis is also performed on the benefits of gain flattening and where it can be omitted. 

\section{EDFA noise figure}
\label{sec:EDFA}

% \hl{FDRO writes this:

% Physical assumptions and definitions of NF

% Relation to ASE 

% }

\begin{figure}[!t]
\centering
 \includegraphics[trim=0 0 0 0cm, width=1.0\linewidth]{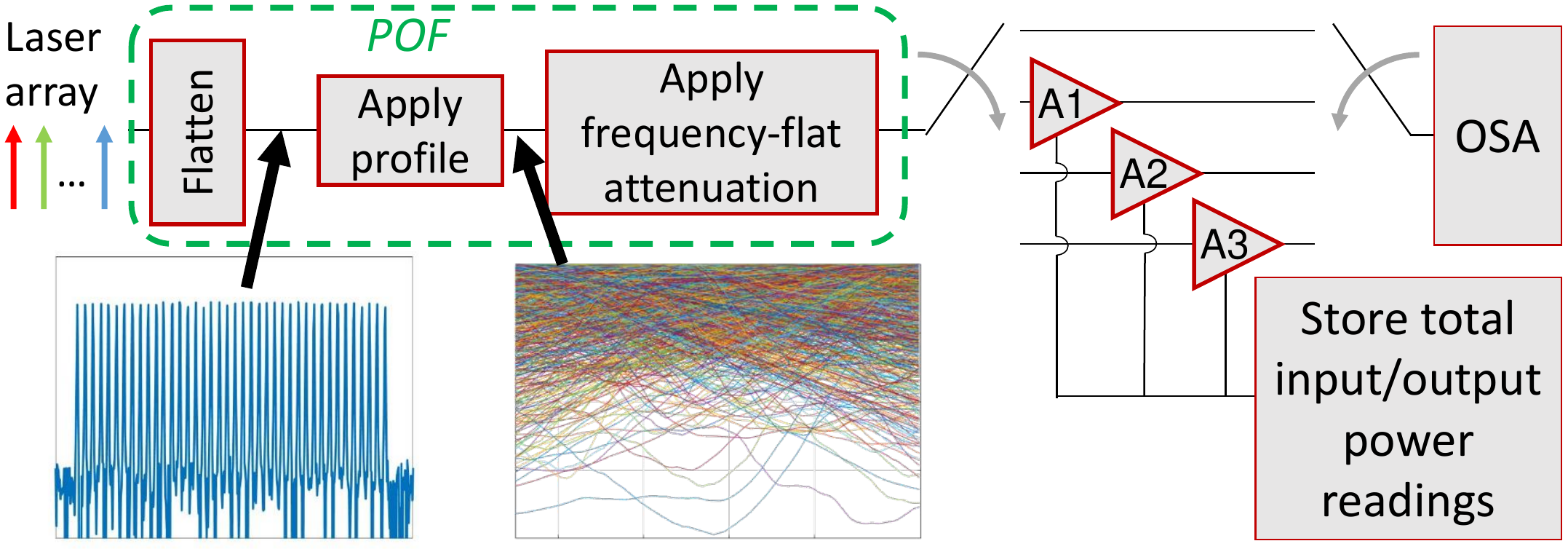}
 \caption{Experimental EDFA characterization setup. A POF is used to 1) flatten the PSD of the laser array; 2) apply the desired profile from the dataset; and 3) attenuate to the desired total power input to the EDFA. The input spectrum and the EDFAs output spectra are recorded sequentially by an OSA.}
 \label{fig:EDFAExpSetup}
%  \vspace{-0.4cm}
\end{figure}

The methodology to extract the gain from input-output PSD relations is similar to that of \cite{YankovJLT}, with the key exception that a laser array is used instead of an input amplified spontaneous emission (ASE) source, allowing the measurement of both gain and noise figure. The characterization setup is given in \fig{EDFAExpSetup}. A laser array with 40 distributed feedback  (DFB) lasers is used as input to a programmable optical filter (POF). The lasers are placed on a 100-GHz ITU grid covering the C-band (from 192.1~THz to 196~THz) and have an OSNR in excess of 50~dB after power equalization~\cite{UiaraOL2021}. The POF specifies the power of each each laser to achieve profiles of varying excursion, smoothness and tilt. The spectrally shaped carriers are sequentially input into the EDFAs under test by using optical switches and the output spectra are captured by an optical spectrum analyzer (OSA). 

The gain profile is extracted by measuring input and output PSDs calibrated to the input and output power readout of the instrument~\cite{YankovJLT}. Only the gain values at the carrier frequencies are considered, leading to 40 gain measurements over the C-band.
To estimate the NF, the ASE noise density of the EDFA at the lasers frequencies is extracted by interpolating the noise PSD in between carriers. From the ASE and gain profiles, the NF at each carrier frequency $\nu$ is extracted as \cite{UiaraOL2021}\cite{derickson1998fiber}
\begin{equation}
    NF = \frac{\rho_{ASE}}{G h \nu} + \frac{1}{G}\,,
\end{equation}
where $\rho_{ASE}$ is the interpolated dual-polarization ASE distribution from the OSA measurements, $G$ is the gain and $h$ is Planck's constant. The high OSNR of the input lasers allows to neglect the input ASE contribution\footnote{The use of an optical source-subtraction method for the NF estimation~\cite{derickson1998fiber} lead to a difference in the final values below our estimated measurement precision.}.

Gain and NF are measured by systematically varying the input PSD, as well as the total input power to the EDFAs  through an additional frequency flat attenuation imposed by the POF. The POF therefore fulfils three different tasks: flattening, spectral shaping and total power tuning, as shown in Fig.~\ref{fig:EDFAExpSetup}. For each profile, the total input/output power combination is varied, so that the total input power  $P_{in}^{tot} \in \left[-9; 9\right]$ dBm and the total output power $P_{out}^{tot} \in \{15, 16, 17, 18\}$ dBm. All input/output profiles are recorded for three different EDFA units of make ``Keopsys, KPS-STD-BT-C-18-SD-111-FA-FA'', which are single stage and do not employ gain-flattening.

\begin{figure}[!t]
\centering
 \includegraphics[trim=20 0 20 0cm, width=1.0\linewidth]{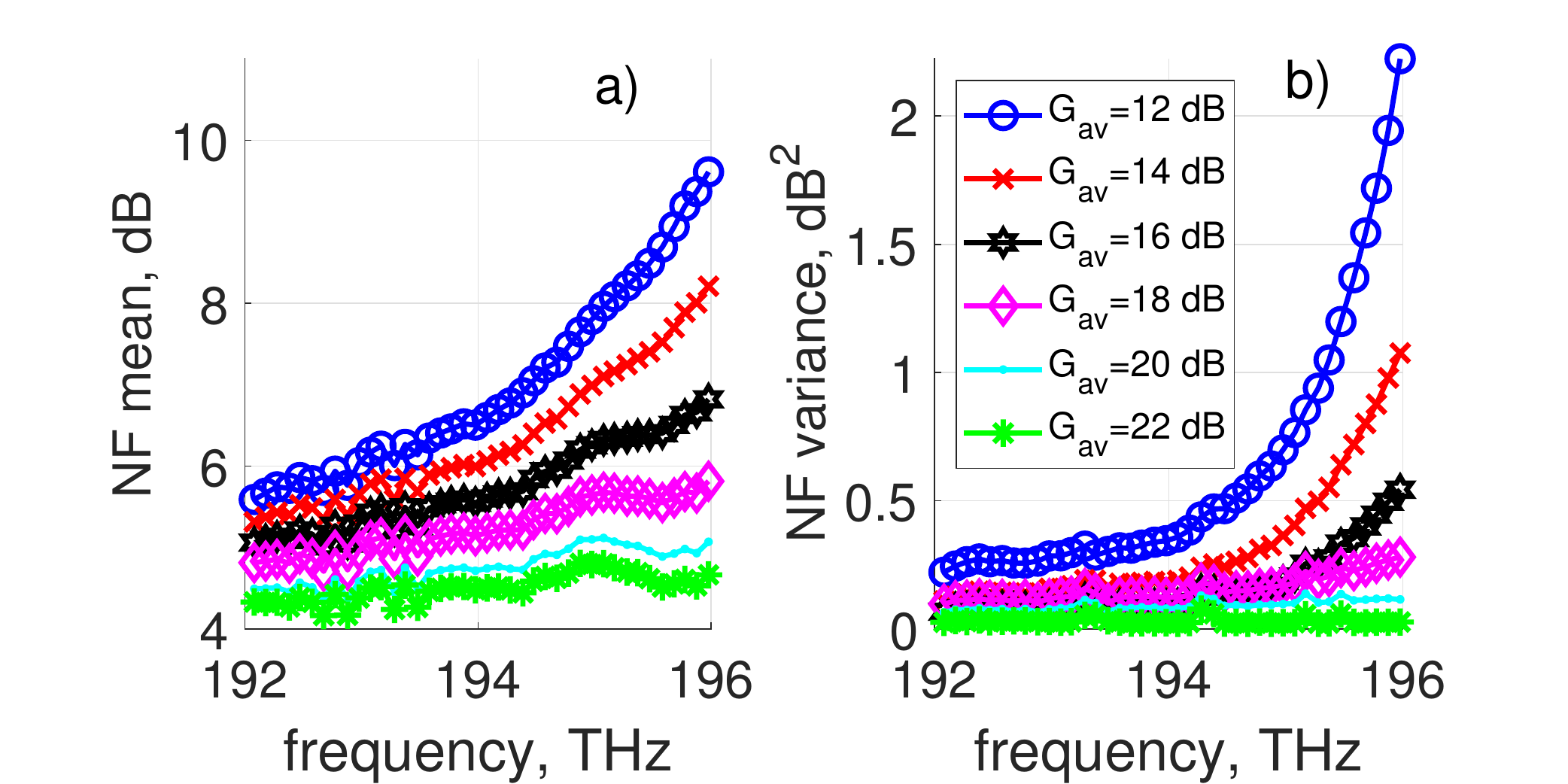}
 \caption{EDFA NF characteristics for variable average gain $G_{av}$. \textbf{a):} average NF profile; \textbf{b):} variance of the NF profile.}
 \label{fig:NFChars}
%  \vspace{-0.4cm}
\end{figure}

The NF characteristics are given in \fig{NFChars} for varying average gain $G_{av} = P_{out}^{tot}-P_{in}^{tot}$. The curves are obtained by averaging the NF profiles in all cases of input/output power combination that fall within a $\pm 1$ dB range of the target $G_{av}$. The NF variance is obtained from the same combinations. The EDFA clearly performs best at high gain, where the lowest NF is observed, as well as the lowest variance, i.e. the NF is least input-dependent in this operating region. For low $G_{av}$, the NF is both higher and more input-dependent, as seen by the variance. 

Details on the measurement profiles and profile measurement procedures are identical to those in \cite{YankovJLT}, where the gain characteristics for the employed EDFAs can also be found. 

\subsection{Machine-learning based modeling}

Theoretical models for frequency-dependent gain profile of an EDFA are typically based on \cite{Saleh}, where the model was first derived. A slight variations is particularly employed in \cite{Perin}. Either implementation requires knowledge of the physical parameters of the EDFA, such as EDF length, radius and loss, doping concentration and pumping configuration. These parameters are not typically provided with the instrument datasheet and are difficult to directly measure experimentally. Furthermore, the physical parameters of the EDF and the doping concentration are always subject to fabrication tolerances. Instead, data-driven approaches can be taken to model the EDFA. 

In this paper, a standard, 2-layer neural network with ReLU activation is employed for the modeling of the NF profile, similarly to what is used for the gain prediction. The network topology and optimization procedure are identical to that of \cite{YankovJLT} with the following differences:
\begin{enumerate}
 \item the input and output dimensions are smaller, corresponding to the 40 carriers;
 \item unlike for the gain, the output for the NF is not normalized since an absolute reference is not available. Instead, the NF is modeled directly. 
\end{enumerate}
To our knowledge, this is the first such model for an EDFA NF with a systematically measured accuracy as shown below.

\begin{figure}[!t]
\centering
 \includegraphics[trim=0 0 0 0cm, width=1.0\linewidth]{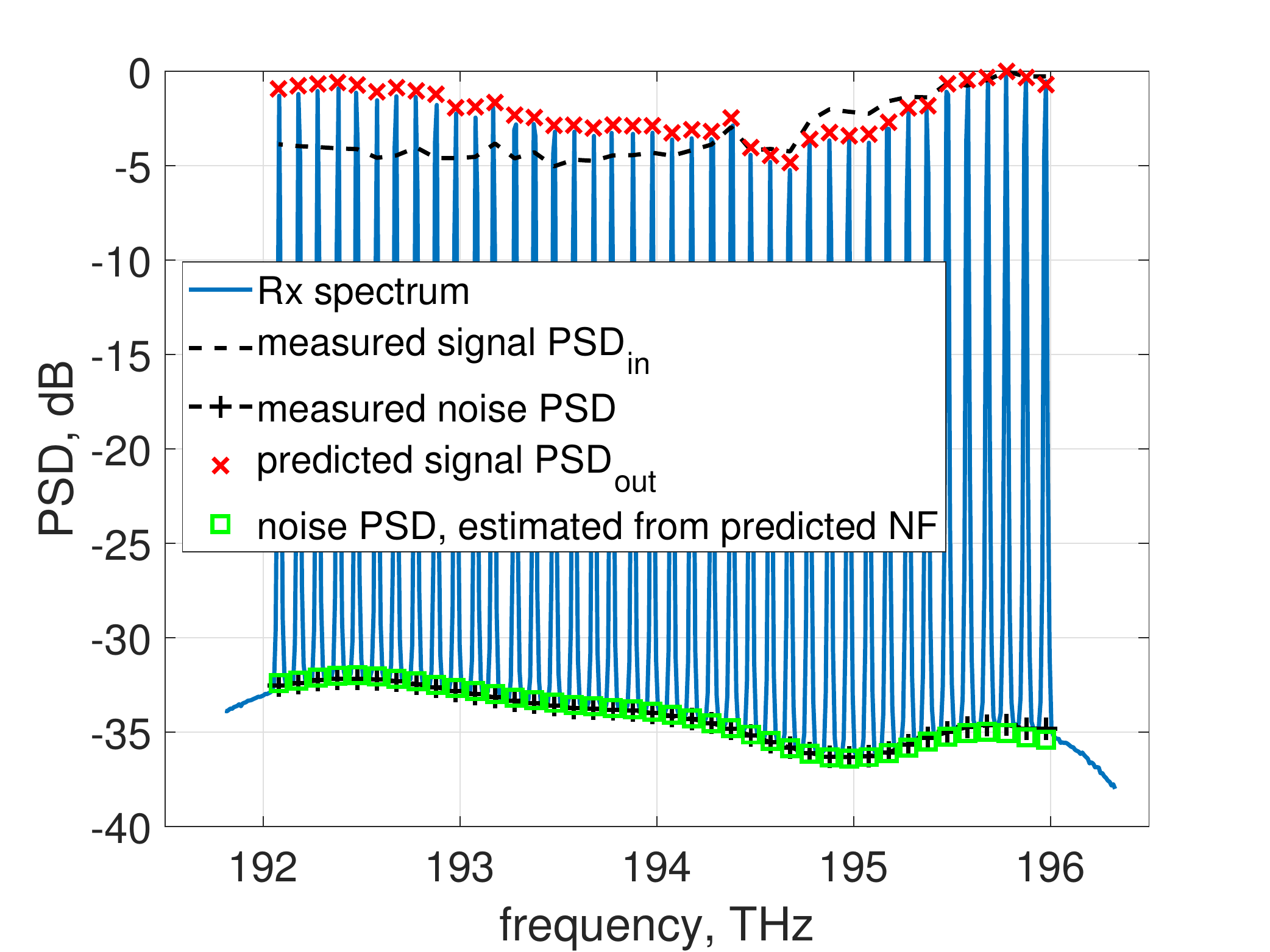}
 \caption{Example of modeling performance. The dashed line represents the envelope of the EDFA input spectrum. The envelope of the output spectrum is accurately predicted (see red crosses), as well as the noise floor: black '+' are interpolated values from the spectrum whereas green squares are estimated from the predicted NF.}
 \label{fig:NFExample}
%  \vspace{-0.4cm}
\end{figure}

\begin{figure*}[!t]
\centering
 \includegraphics[trim=70 0 70 0cm, width=1.0\linewidth]{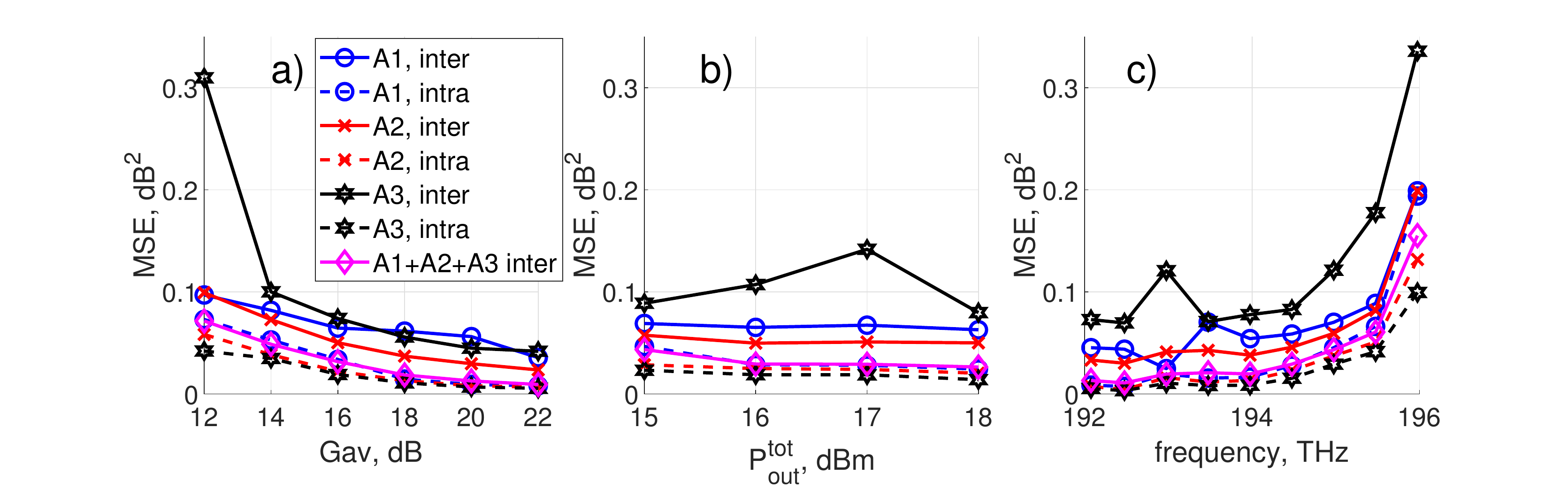}
 \caption{Summary of the NF modeling performance. Intra-MSE means trained and tested on the same EDFA (for different profiles). Inter-MSE means trained on one EDFA and tested on the other two (for different profiles). Intra- and inter-EDFA MSE as a function of \textbf{a):} the average gain for all studied amplifiers;  \textbf{b):} the total output power, averaged across average gains;  \textbf{c):} the channel frequency, demonstrating that the MSE is dominated by the performance at high frequencies.}
 \label{fig:EDFA_MSE_summary}
%  \vspace{-0.4cm}
\end{figure*}

A NN with the new topology is also trained for the gain, and is validated to perform identically to that of \cite{YankovJLT}, where a shaped, continuous ASE source was employed. The newly trained gain NN for the 40-carrier case together with the NF NN compose the complete EDFA model applied in the rest of the paper. 

An example of the modeling performance of the NF is given in \fig{NFExample} for a random input $PSD_{in}$. The normalized received spectrum is shown together with the predicted signal $PSD_{out}$ and the predicted ASE noise density, as extracted from the predicted NF. A summary of the modeling performance is given in \fig{EDFA_MSE_summary} in terms of \textit{intra-} mean squared error (MSE) (training and testing data from the same EDFA) and \textit{inter-}MSE (training on a single EDFA and testing on the other two). The modeling performance is strongly dependent on the average gain and is worst at low $G_{av}$, as can be expected from the increased variance in that operating region. Nevertheless, an intra-MSE below $0.07 \mbox{dB}^2$ at low $G_{av}$ is achieved, reduced to below $0.01 \mbox{dB}^2$ at high $G_{av}$. Similar performance is seen for the case of training and testing jointly on all EDFAs (indicated as 'A1+A2+A3 inter' in the figure legend). Degradation of up to $0.03 \mbox{dB}^2$ is seen when the model is trained and tested on different EDFA units (except for the outlier point of the A3 inter-MSE at $G_{av}=12$ dB, which is due to too few training points available for that combination). The modeling performance is relatively constant as a function of output power (\fig{EDFA_MSE_summary}b)), with a slight disadvantage when training on EDFA A3. As seen from \fig{EDFA_MSE_summary}c), the MSE is dominated by the performance at high frequencies, similar to the case of modeling the gain \cite{YankovJLT}. 

% \begin{mdframed}[hidealllines=true,innerleftmargin=0pt,innerrightmargin=0pt,innertopmargin=0pt,innerbottommargin=0pt,backgroundcolor=yellow]
For comparison, the standard deviation of the noise model cited in \cite{JuhnoECOC2020} is $0.24$ (not stated if intra- or inter-modeling performance), which corresponds to an MSE of $0.06$, and for the gain is about $0.33$ (corresponding to an MSE of $0.11$). The standard deviation across all studied gains and output powers for the proposed model is for the NF $0.14$ dB and $0.25$ dB for the intra- and inter-modeling performance repsectively, and for the gain $0.14$ dB and $0.2$ dB for the intra- and inter-modeling performance, respectively. If the combinations of simultaneous low gain ($< 15 dB$) and low output power ($P_{out}^{tot}=15$ dBm), which are outside the nominal operation range of the studied EDFA are excluded, the standard deviations are $0.1$, $0.2$, $0.1$ and $0.17$ dB for the four cases, respectively. 

\begin{figure}[!t]
% \begin{mdframed}[hidealllines=true,innerleftmargin=0pt,innerrightmargin=0pt,innertopmargin=0pt,innerbottommargin=0pt,backgroundcolor=yellow]
\centering
 \includegraphics[trim=0 0 0 0cm, width=1.0\linewidth]{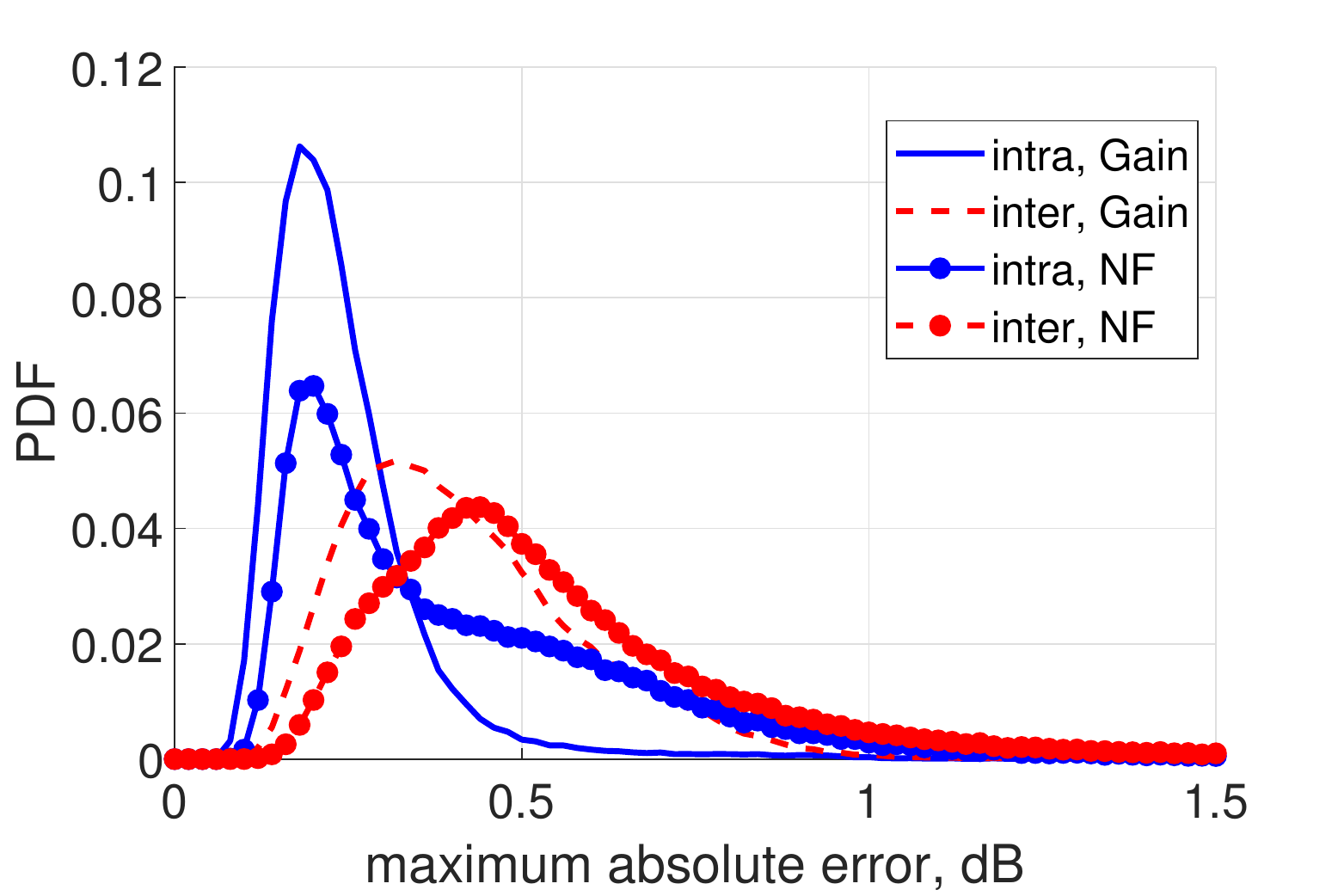}
 \caption{PDF of the MAE for the inter- and intra-model performance of the gain and NF models.}
 \label{fig:MAE}
%  \end{mdframed}
%  \vspace{-0.4cm}
\end{figure}

The modeling performance is further analyzed in terms of maximum absolute error (MAE), defined as
\begin{gather}
 MAE_{G} = \max_n |PSD_{out}(n)-\hat{PSD}_{out}(n)|, \nonumber \\
 \label{eq:MAE}
 MAE_{NF} = \max_n |NF(n)-\hat{NF}(n))|,
\end{gather}
 for the gain and NF, respectively. In \eq{MAE}, $PSD_{out}(n)$ and $\hat{PSD}_{out}(n)$ are the measured and predicted output profiles for frequency index $n$, respectively, and $NF(n)$ and $\hat{NF}(n)$ are the NFs estimated from the measured ASE and predicted by the model, respectively. The probability distribution function (PDF) of the intra- and inter-model MAE is given in \fig{MAE} for the gain and NF models. The NF exhibits a higher maximum and wider spread, which we attribute to the higher uncertainty in measuring the noise floor which ultimately drives the accuracy of the ASE measurements and therefore the model. It is noted that the high errors are mainly attributed to the low-gain, low-power regime described earlier and to input profiles of relatively extreme shapes in terms of tilt and excursion. Such cases and are not detrimental to the modeling of the average few-span link, which will be the focus in the next sections. This has also been demonstrated in \cite{JuhnoECOC2020}\cite{YankovJLT}. 
% \end{mdframed}

The relatively low inter-EDFA MSE allows for a multi-span system to be modeled using training data from a single unit. In practical networks, a model can thus be built for the EDFA of each make that is present in a network. Such a collection of models can then be used to model the entire network with minimal pre-characterization required, including for networks which are already deployed. Alternatively, if extreme accuracy is required, a model can be trained for each unit that would be deployed in the network, at the cost of significant pre-characterization effort prior to deployment. 

\section{Multi-span system model and optimization}

\begin{figure}[!t]
\centering
 \includegraphics[trim=0 0 0 0cm, width=1.0\linewidth]{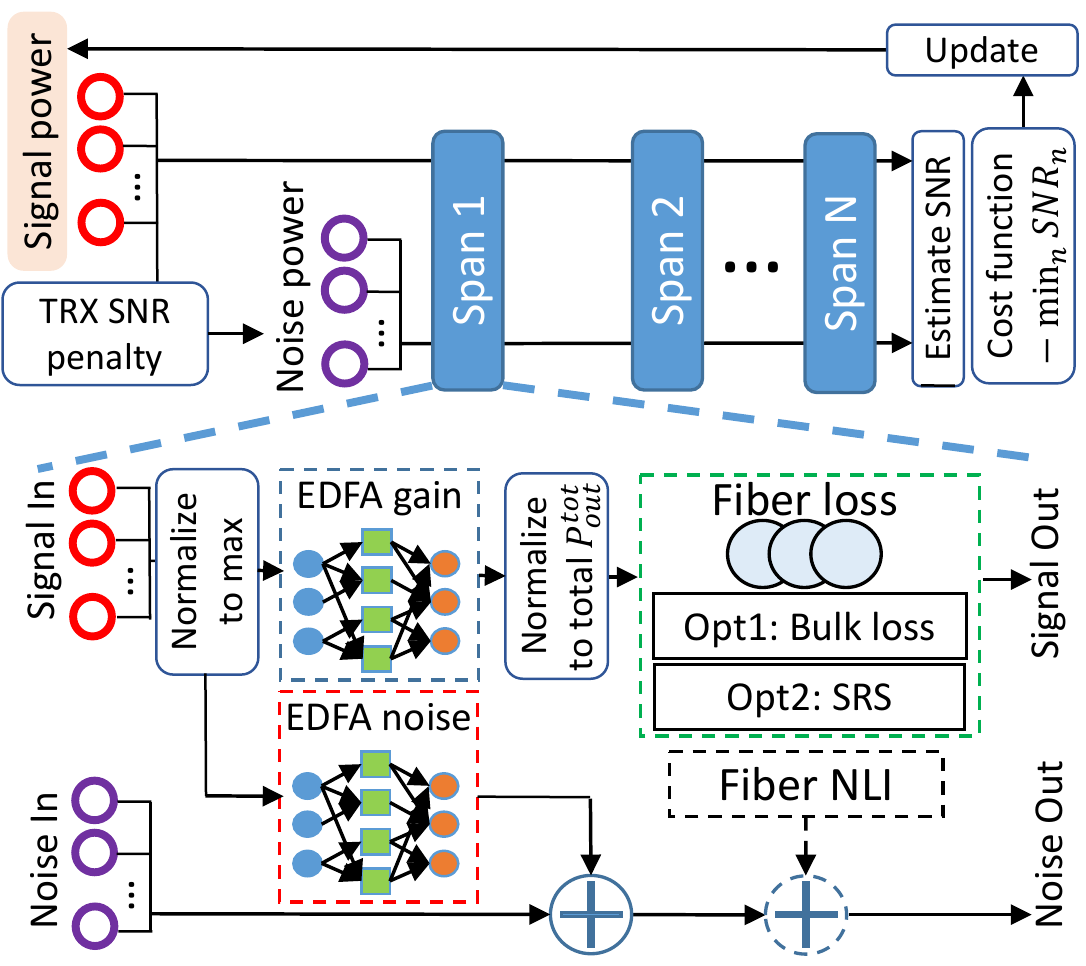}
 \caption{\textbf{Top:} Multi-span system model and optimization. \textbf{Bottom:} Single span model with EDFA and fiber components. The fiber model has SRS and NLI as options to be included.}
 \label{fig:optimization_model}
%  \vspace{-0.4cm}
\end{figure}

The EDFA gain NN model, the NF NN model and a GN model~\cite{Poggiolini} for the fiber (including the SRS contributions \cite{Roberts}\cite{Semrau}) are combined to build an arbitrary configuration of number of spans, length per span and launch power, as shown in~\fig{optimization_model}. The assumed configuration employs an EDFA as a first component, which is fed by the input PSD. The gain and NF model are used to predict the PSD at the input of the fiber and the current ASE contribution to the total noise in the system. At the first span, the EDFA noise contribution is added to the implementation penalty, which can be obtained experimentally for the transmitter and receiver under test. For every subsequent span, the EDFA noise contributions accumulate. The EDFA gain model is trained such that its input and output are normalized to $\max_n PSD(n) = 0$, which we found improved the accuracy of the model. Before applying the fiber model, the PSD is thus normalized to the desired total launch power. 

The GGN fiber model can be parametrized in various ways depending on the complexity and desired accuracy. For example, a bulk-loss model is employed in \cite{JuhnoECOC2020}\cite{JuhnoChoJLT}, a bulk-loss with Kerr nonlinear interference (NLI) in \cite{RobertsOld}, and a complete model with SRS and Kerr nonlinearities in \cite{Roberts}\cite{Semrau}\cite{TorinoECOC2020}. If Kerr NLI is applied, its contribution is added to the total accumulated noise. After all spans have been modeled, the system estimates both the total PSD, as well as the total accumulated noise, from which the SNR is estimated. In this work, a flat attenuation profile of the fiber is considered. However, a non-flat profile is straight-forward to include. This is especially relevant for e.g. ultra-wide band systems employing transmission beyond the C-band. Depending on the context (e.g. optimization vs. prediction), different parametrizations are applied in this work, as detailed below. 

The SRS model in this paper directly follows \cite{YankovJLT}. The Kerr-induced cross-phase and self-phase modulation (XPM and SPM, respectively) NLI are modeled with the basic GN model \cite{Poggiolini}. When the fiber input is Gaussian distributed, the GN model is identical to the more general enhanced GN model \cite{Carena} which has been shown accurate even at short distances. Gaussian distribution in this paper is closely approached by constellation shaping (to be discussed below). The standard GN model is thus chosen due to its simplicity.

\begin{figure*}[!t]
\centering
 \includegraphics[trim=0 0 0 0cm, width=1.0\linewidth]{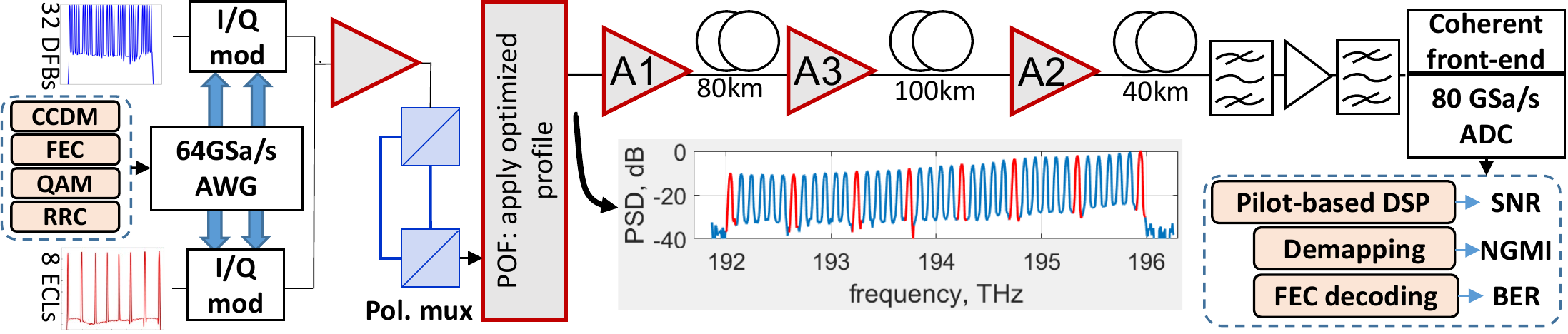}
 \caption{System experiment block diagram. }
 \label{fig:exp_setup_system1}
%  \vspace{-0.4cm}
\end{figure*}

All functionalities described above, namely 1) the EDFA gain model; 2) EDFA noise model; 3) SRS-induced power evolution; and 4) Kerr nonlinearities are differentiable w.r.t. the input PSD of each corresponding system component. A cost function of the form $Cost=-\min_n SNR(n)$, where $SNR(n)$ is the SNR at the $n$-th carrier can then be applied in order to maximize and equalize the SNR profile of the system. Alternatively, a cost function of the form $Cost=-\sum_n\log_2(1+SNR(n))$ can be applied in order to maximize the total throughput of the system \cite{JuhnoECOC2020}\cite{JuhnoChoJLT}. The former cost function is chosen as target in this paper to exemplify the optimization method experimentally.  

For the rest of the paper, all system models used for PSD optimization employ an EDFA gain and NF models which are trained on data from \textit{amplifier A1 only} in order to demonstrate the generalization capability of the ML model. 

The following optimization strategies are considered in this paper:
\begin{enumerate}
 \item Flattening the received power profile \cite{YankovJLT}.
 \item Flattening the received SNR by neglecting the SRS and Kerr NLI. This is similar to the solution in \cite{JuhnoChoJLT} for optimization of the OSNR, with the difference that in this paper the implementation penalties are taken into account.
 \item Flattening the received SNR by including the SRS and Kerr NLI in the model. 
\end{enumerate}
The performance of a flat input profile is also measured for reference.

\section{Experimental demonstration}

The experimental setup for validation of the cascaded model and demonstrating the SNR optimization is given in \fig{exp_setup_system1}. In order to ensure that the GGN model-predicted XPM and SPM NLI contributions are accurate, 256QAM is employed with probabilistic shaping using a Maxwell-Boltzmann distribution with an entropy of 6 bits/symbol, which is a near-optimal choice for maximizing the normalized generalized mutual information (NGMI), and thus the error-free data rate per channel at the target SNR range of $\left[15; 18\right]$ dB. A constant composition distribution matching (CCDM)-based probabilistic amplitude shaping (PAS) is applied \cite{PAS}. The forward error correction (FEC) code employed is the DVBS-2 low-density parity check (LDPC) code with an overhead of 33\%. A square-root raised cosine pulse shaping is applied with a roll-off factor of 0.01, and the signal is sent to an arbitrary waveform generator (AWG). Due to experiment duration restrictions, the SNR performance of 8 channels is targeted for measurements. In order to improve the performance and frequency stability at least on those channels, eight of the DFB lasers used for the EDFA characterization are replaced by external cavity lasers (ECLs, 10-kHz of linewidth). The ECLs are placed uniformly within the bandwidth (inset in \fig{exp_setup_system1}). Independent data are used to modulate the DFBs and the ECLs at 32 GBd. The two modulated channel banks (ECLs and DFBs) are combined and amplified. Polarization emulation is applied using a standard delay-and-add technique and the 40-channel signal is sent to the POF for applying the desired power profile. The POF is controlled with a feed-back loop that ensures a maximum of $0.2$ dB of error between the desired and the measured power profile at the input of the system.

\begin{figure}[!t]
\centering
 \includegraphics[trim=0 0 0 0cm, width=1.0\linewidth]{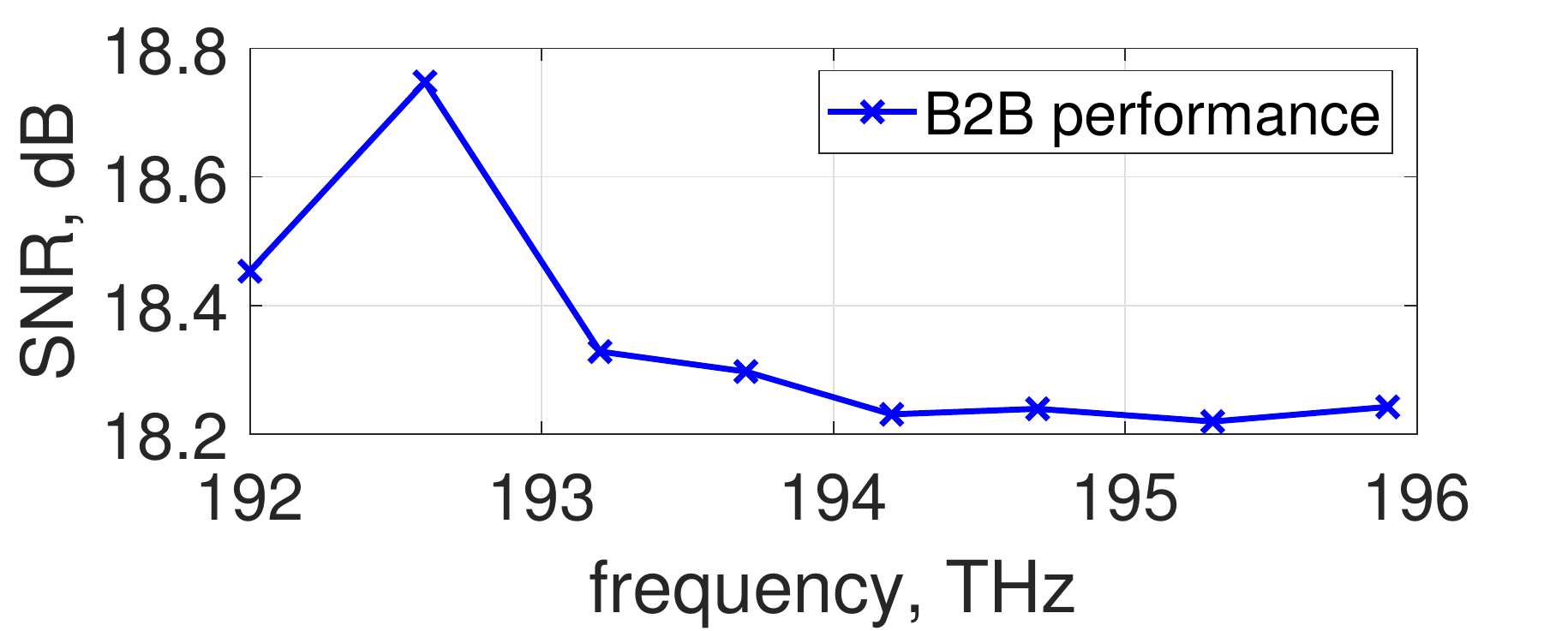}
 \caption{B2B performance of the ECL carriers.}
 \label{fig:SNR_vs_f_B2B}
%  \vspace{-0.4cm}
\end{figure}

The experimental penalty at the ECL positions can be derived from the back-to-back (B2B) performance for a flat input profile, which is given in \fig{SNR_vs_f_B2B}. Up to $0.6$ dB of difference can be seen, which as mentioned above is taken into account when predicting the system output SNR profile and optimizing it. The variations are mainly attributed to the impact of the receiver pre-amplifier and a varying transmitter OSNR at the different channels, which in originates in the noise floor of the EDFA prior to the POF. 

\begin{figure*}[!t]
\centering
\subfigure[$PSD_{in}$, \textbf{A1-80km-A3-100km-A2-40km}]{
 \includegraphics[trim=0 0 0 0cm, width=0.31\linewidth]{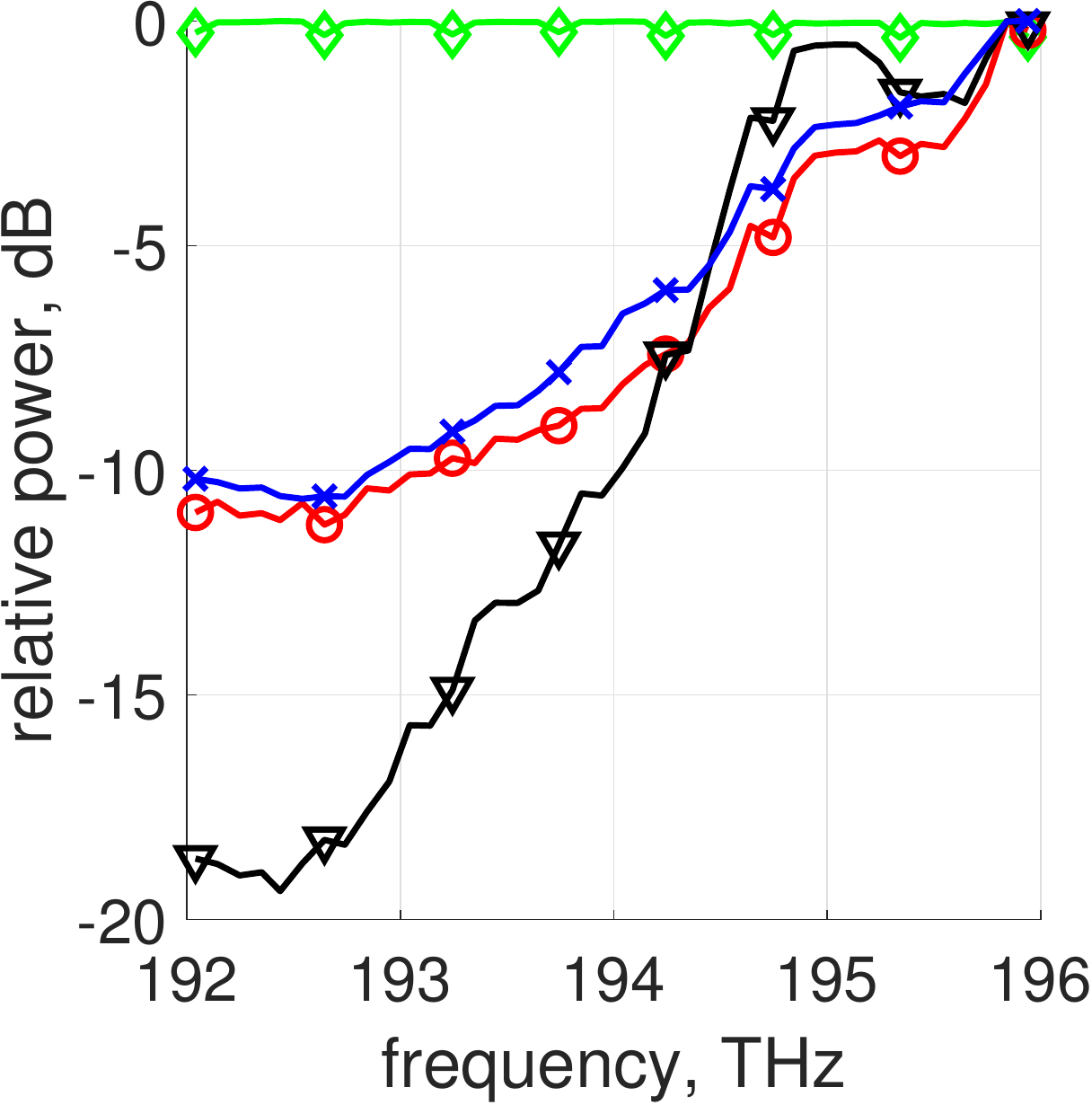}}
 \subfigure[$PSD_{out}$, \textbf{A1-80km-A3-100km-A2-40km}]{
 \includegraphics[trim=0 0 0 0cm, width=0.31\linewidth]{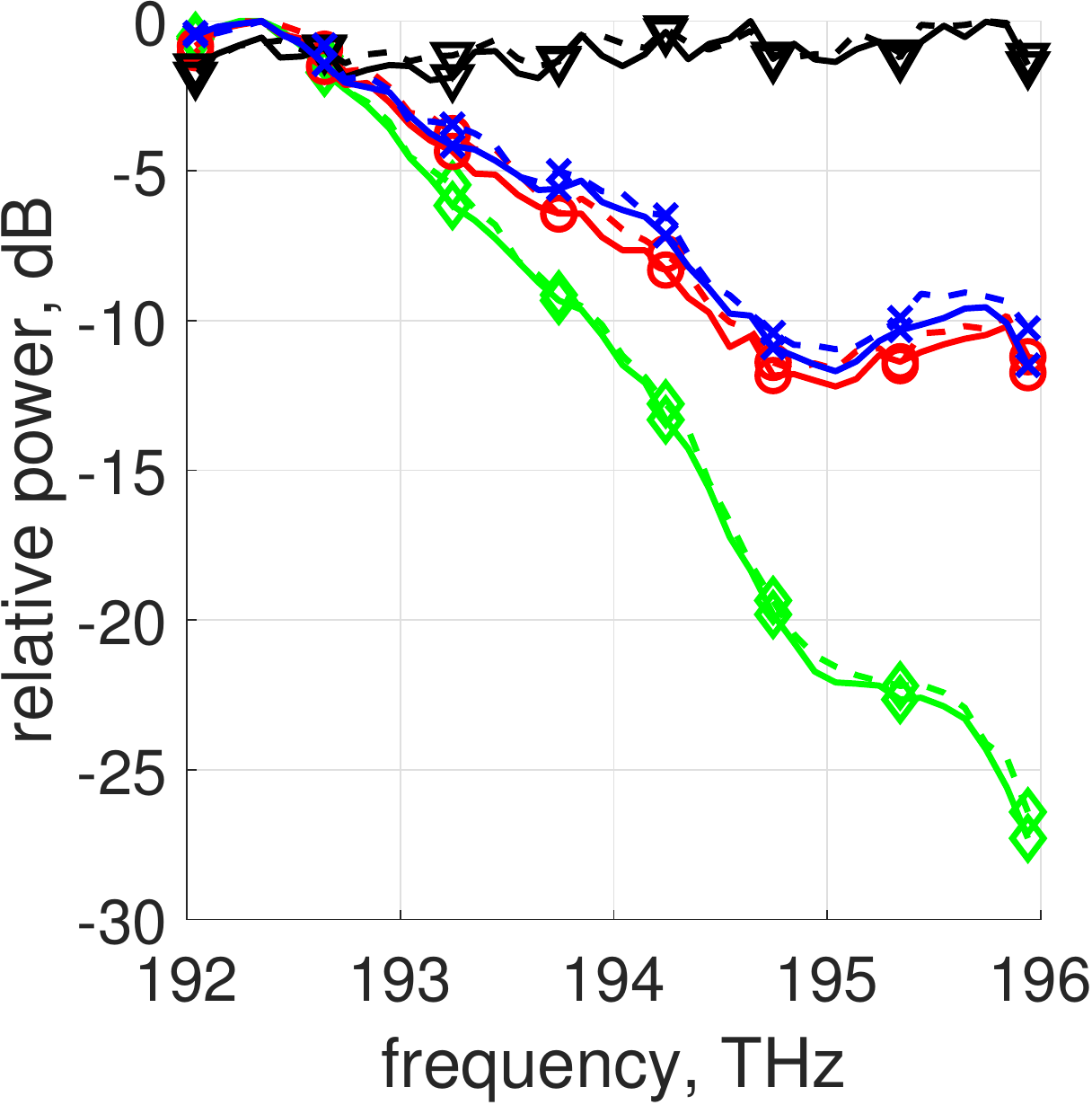}}
 \subfigure[SNR, \textbf{A1-80km-A3-100km-A2-40km}]{
 \includegraphics[trim=0 0 0 0cm, width=0.31\linewidth]{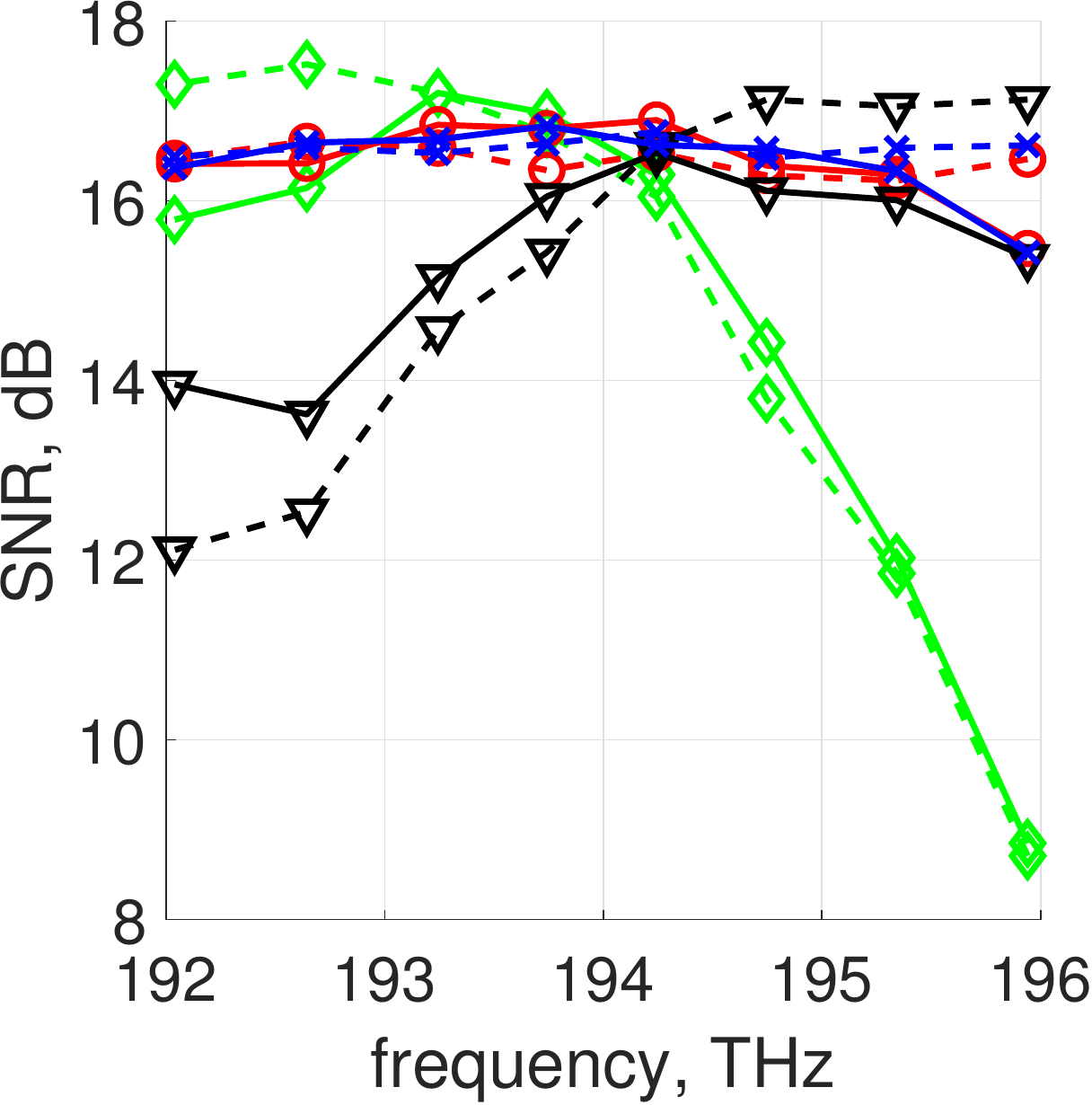}}
 \\
 \subfigure[$PSD_{in}$, \textbf{A1-120km-A2-100km-A3-80km}]{
 \includegraphics[trim=0 0 0 0cm, width=0.31\linewidth]{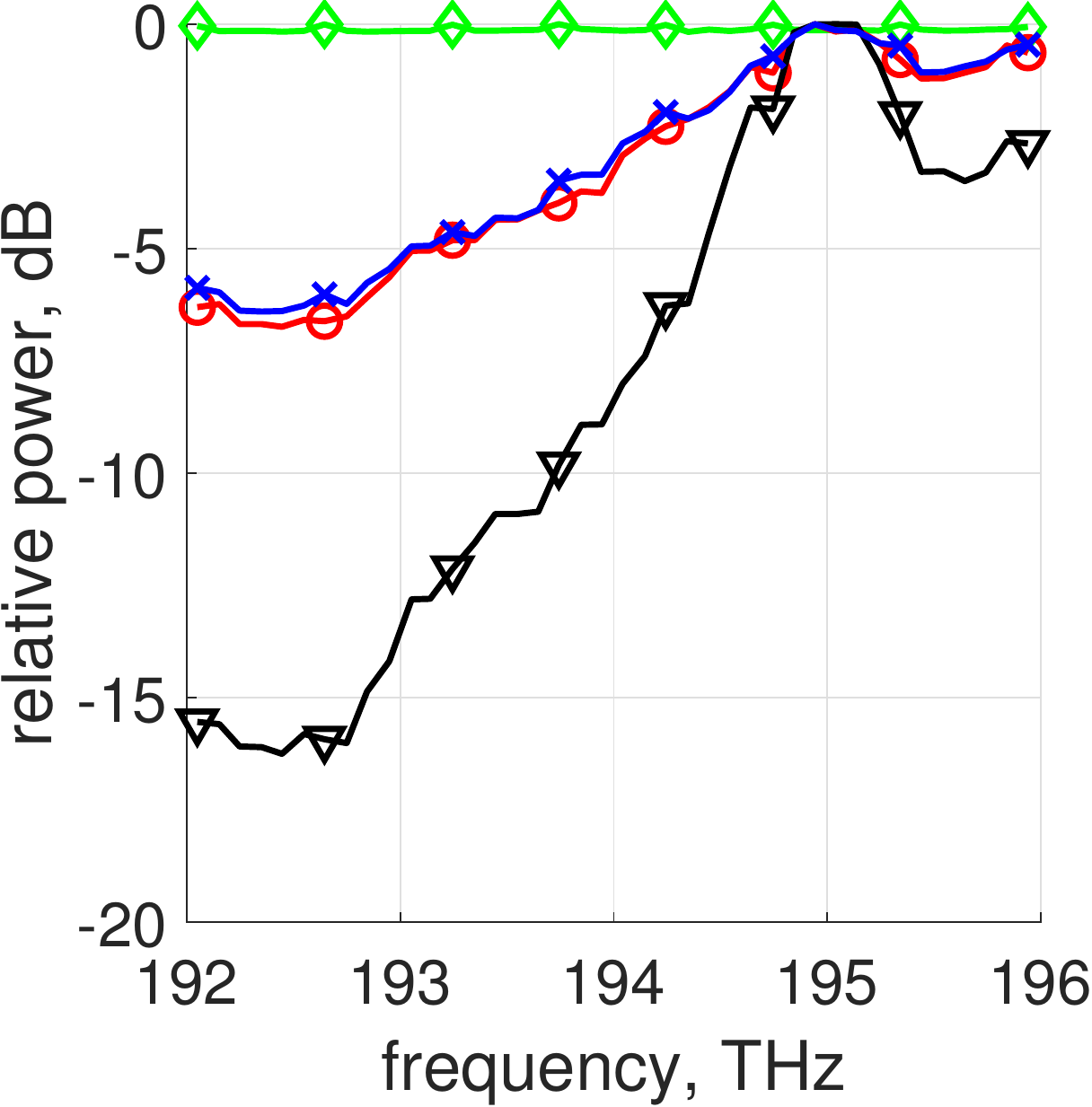}}
 \subfigure[$PSD_{out}$, \textbf{A1-120km-A2-100km-A3-80km}]{
 \includegraphics[trim=0 0 0 0cm, width=0.31\linewidth]{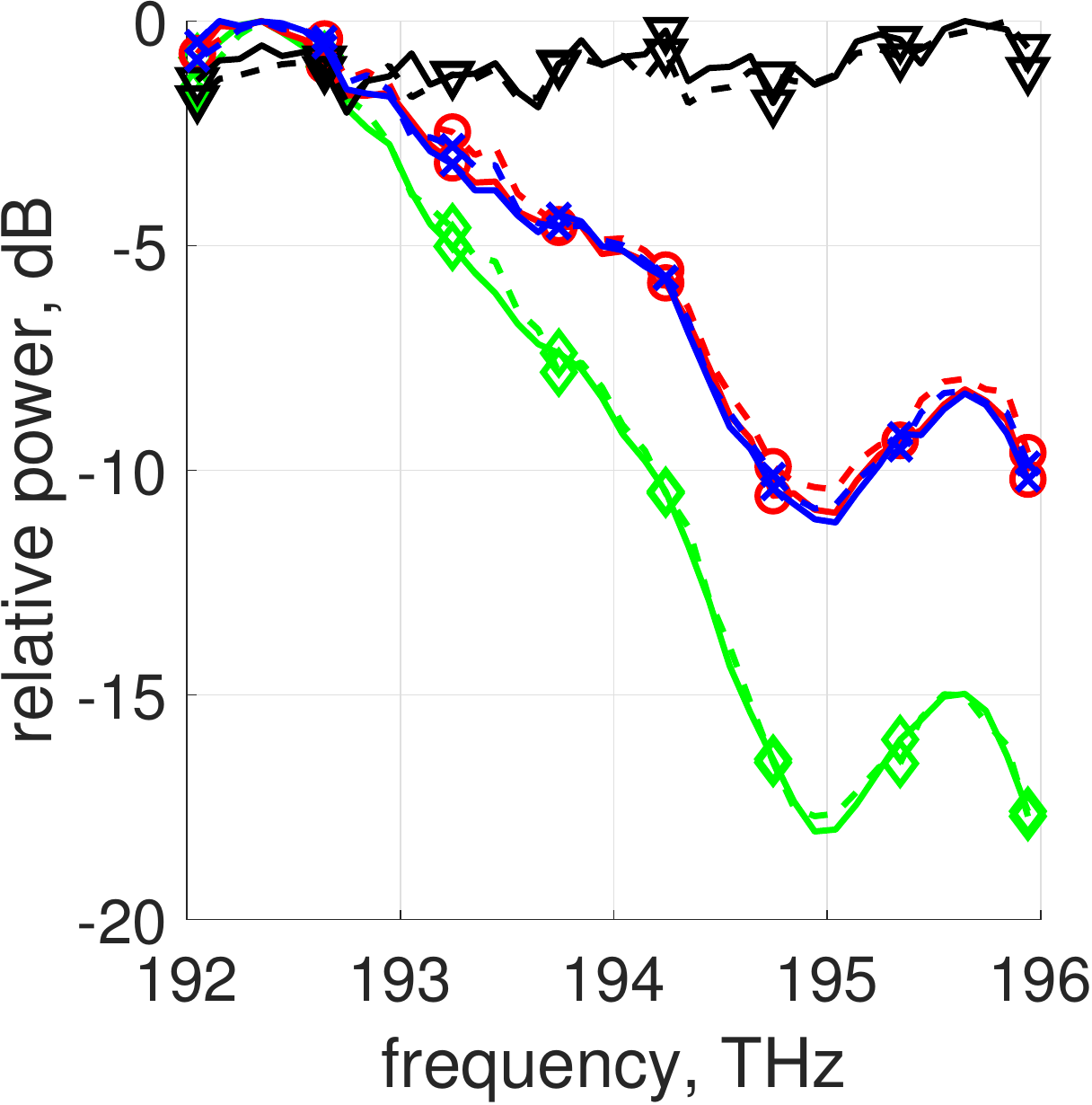}}
 \subfigure[SNR, \textbf{A1-120km-A2-100km-A3-80km}]{
 \includegraphics[trim=0 0 0 0cm, width=0.31\linewidth]{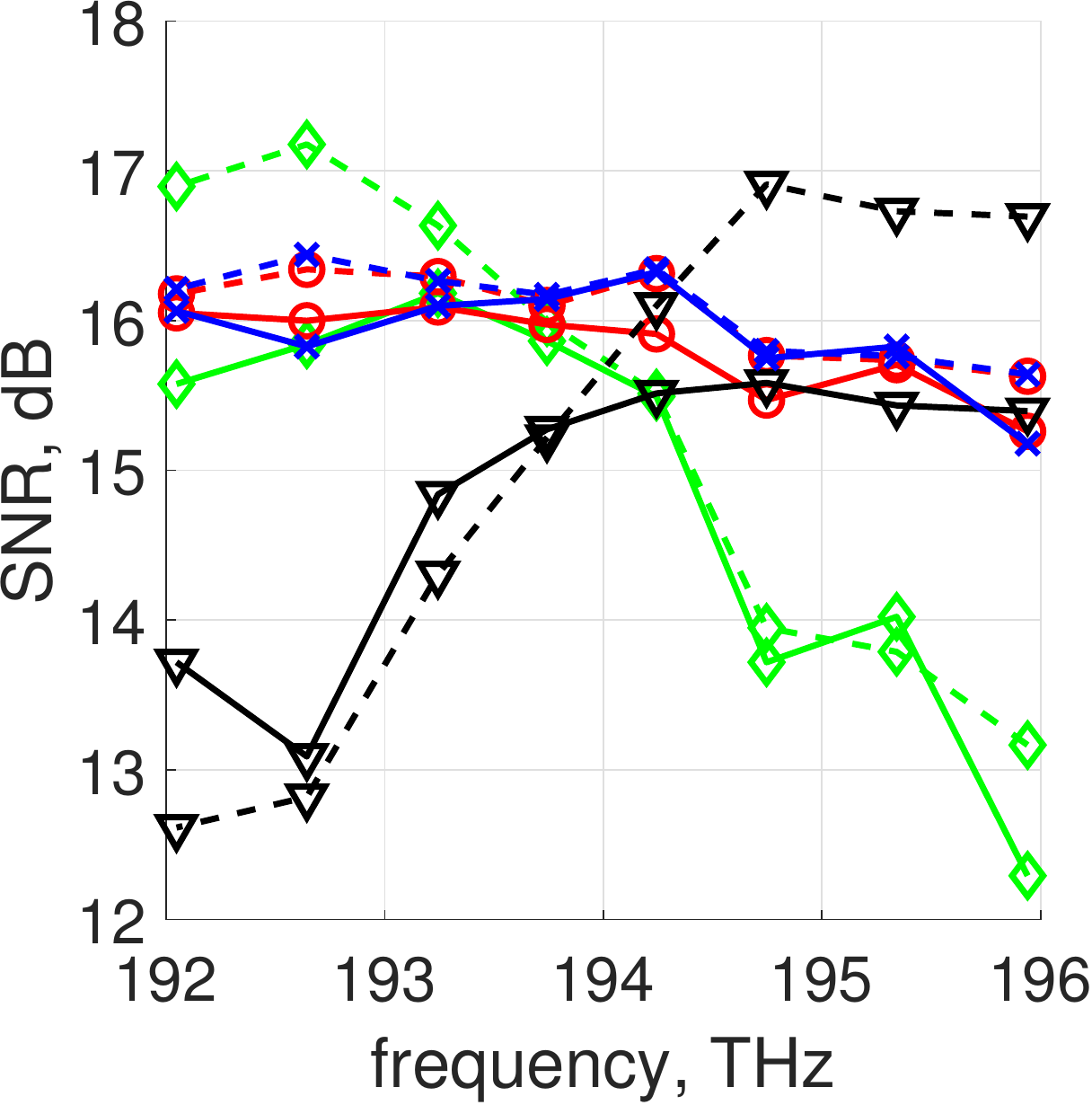}}
 \\
 \subfigure{
 \includegraphics[trim=0 0 0 0cm, width=1.0\linewidth]{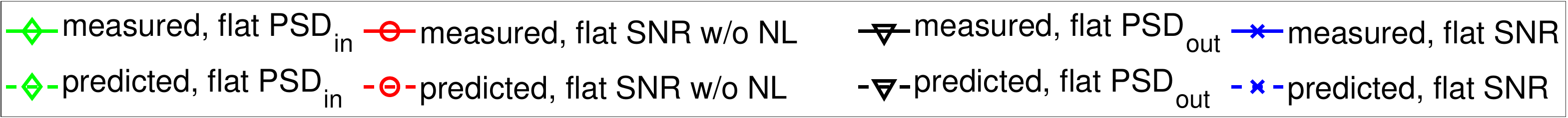}}
 \caption{Experimental results for two arbitrarily chosen configurations of 3 spans. \textbf{a,d):} Optimized power profiles measured at the system input. \textbf{b,e):} Corresponding received power profiles together with model predictions. \textbf{c,f):} Corresponding received SNR at the ECLs positions together with model predictions.}
 \label{fig:system_perf}
%  \vspace{-0.4cm}
\end{figure*}

The system under test follows, exemplified in \fig{exp_setup_system1} with a 3-span system with fiber lengths of 80~km, 100~km and 40~km, amplified by EDFAs A1, A3 and A2, respectively. The maximum total output power of the employed EDFAs is $18$~dBm, which was found to be superior than lower ones in terms of system performance for all studied optimization strategies. It was therefore chosen as the launch power into each span. The total power input to the first EDFA was arbitrarily chosen -2 dBm, resulting in $G_{av}=20$ dB for the first EDFA. The gain values for the other two EDFAs were measured as $G_{av}=15.6$ dB and $G_{av}=20.6$ dB, respectively, and compensate for the all fiber and connector losses between EDFAs. An example spectrum of an input signal, optimized for this system by including all components impairments is given in the inset of \fig{exp_setup_system1}. At the system end, the WDM signal is band-pass filtered (BPF) to select the channel under test, amplified to the power required by the coherent receiver, filtered again and sent to the coherent receiver front-end. The purpose of the first filter is to mitigate the above-mentioned frequency dependency of the receiver EDFA, whereas the second BPF is employed so that the receiver photodiodes are not saturated with noise. The signal is downconverted with a 10-kHz ECL acting as local oscillator and analog to digital converted (ADC) using an 80~GSa/s digital storage oscilloscope. Offline processing is then applied to measure the performance in terms of SNR, NGMI and bit error rate (BER). The processing is identical to that of \cite{YankovOPC}.

\section{Results}

The results for the system labeled \textbf{A1-80km-A3-100km-A2-40km}, indicating the order of amplifiers and fiber spools, is given in the top 3 subfigures of \fig{system_perf}. In \fig{system_perf}a), the normalized optimized input profiles are shown. The profiles targeting flat SNR are similar, and exhibit an excursion of $\approx10$~dB. Flattening the system output PSD requires an excursion of $\approx$~20 dB at the input of this system, which was also demonstrated in \cite{YankovJLT}. The measured system output PSDs are given in \fig{system_perf}b), together with the corresponding predicted PSDs from the fiber model that includes SRS. An excellent correspondence is observed between the experimental measurements and the model predictions for all profiles, which validates the flat fiber attenuation profile assumption taken during modeling for this system. Finally, the SNR at the ECL positions is given in \fig{system_perf}c). 

A flat input profile results in a strong SNR excursion of $\approx8$~dB, with the worst channel exhibiting an SNR $\approx 9$~dB. The channels at high frequencies exhibit worse performance due to the combined effect of the strong EDFA gain and NF tilt and the SRS resulting in a very poor OSNR. Flattening the received PSD somewhat mitigates the excursion to $\approx2.5$ dB, with the worse channels having a lower frequency, i.e., overcompensation of SNR is observed. The solutions which target the SNR are much more balanced and result in an improvement of the minimum SNR to $\approx15.2$ dB with an excursion of $\approx 1.2$ dB. 

We see that for the flattened SNR profiles, up to 1 dB of SNR prediction error is achieved, which is worst at the higher frequencies. This is explained by the EDFA modeling performance at high frequencies seen in \fig{EDFA_MSE_summary}, and together with the other implementation uncertainties is the main limitation to further improving the performance of the worst performing channels and further flattening the SNR profile. The SNR prediction for the flat input profile and the profile that flattens the output PSD is worse. This can be explained by the fact that strong PSD excursions are experienced at the input of multiple EDFAs in the system. Such excursions are at the boundaries of the modeling capabilities of the EDFA model, both for gain and NF, and as demonstrated in \cite[Fig. 7]{YankovJLT} experience increased modeling error, which is eventually transferred to SNR error.

In order to demonstrate the reconfigurability of the proposed SNR equalization method, the performance of a system labeled \textbf{A1-120km-A2-100km-A3-80km} with corresponding span lengths of 120 km, 100 km and 80 km, respectively is given in the bottom 3 subfigures of \fig{system_perf}. The gain values for the three EDFAs in this case were measured as $20$, $23.5$ and $20.6$ dB, respectively. The qualitative analysis is identical to that of \textbf{A1-80km-A3-100km-A2-40km} above. 

No errors were observed after FEC decoding for either system for all channels with the SNR-flattening profiles, resulting in error-free transmission with a data rate of 256~Gbps per channel.

% \begin{mdframed}[hidealllines=true,innerleftmargin=0pt,innerrightmargin=0pt,innertopmargin=0pt,innerbottommargin=0pt,backgroundcolor=yellow]
\section{Network performance}
\begin{figure}[!t]
% \begin{mdframed}[hidealllines=true,innerleftmargin=0pt,innerrightmargin=0pt,innertopmargin=0pt,innerbottommargin=0pt,backgroundcolor=yellow]
\centering
 \includegraphics[trim=0 0 0 0cm, width=1.0\linewidth]{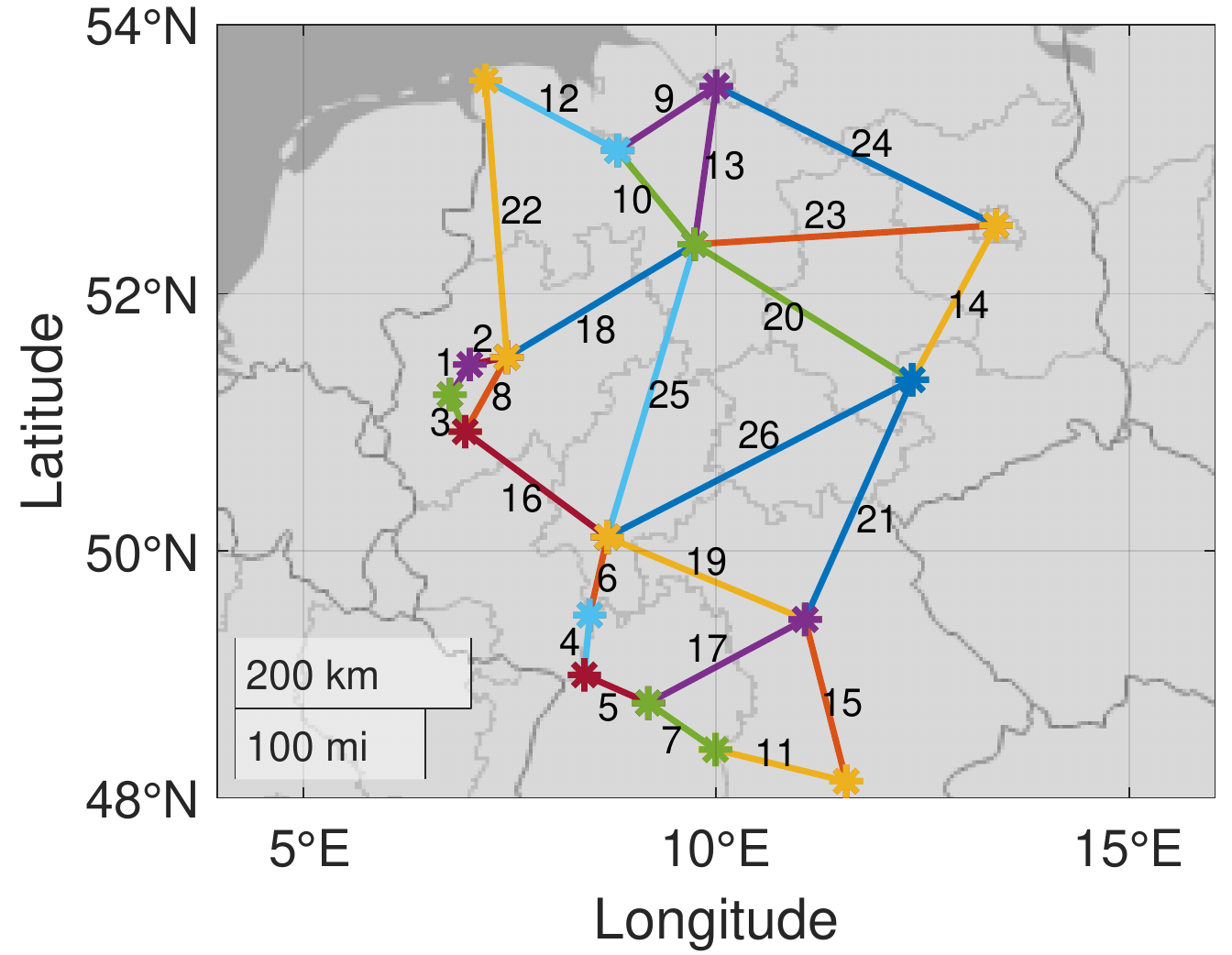}
 \caption{Topology of the German core network considered in this work, with all links annotated by indexes.}
 \label{fig:GermanTopology}
% \end{mdframed} 
%  \vspace{-0.4cm}
\end{figure}

In order to further exemplify a GFF-free operation with the studied EDFA, the German core network topology is considered, given in \fig{GermanTopology}. The network consists of 17 nodes and 26 links in total, with distances between 36 and 353 km. In such core transport networks, fully loaded WDM operation can be assumed with re-routable optical add-drop multiplexer at the nodes with optical-electrical-optical conversion taking place \cite{Betker:14}. Optimizing the performance for each link is therefore relevant for maximizing the network performance overall. Each link is then simulated using the cascade model in \fig{optimization_model} and optimized as above. 

% \end{mdframed}

% \begin{mdframed}[hidealllines=true,innerleftmargin=0pt,innerrightmargin=0pt,innertopmargin=0pt,innerbottommargin=0pt,backgroundcolor=yellow]
For clarity, the links are numbered in ascending order of their length. For reference, a fixed 75 km span length was assumed for all links in \cite{Vittorio}. Here, a random span length between 36 and 100 km is chosen to exemplify the reconfigurability of the proposed system model in terms of amplifier gain.  The randomly generated spans are given in Table~\ref{tbl:spanLenghtDistribution}. For completeness, the case of \textit{ideal} gain flattening at every amplification stage is also considered. The ideal GFF stage is modeled as an extra amplifier with a NF of 3 dB and a filter which flattens the output power profile. The gain of the extra amplifier is set to the minimum required to boost the channel with lowest power to the desired common level, i.e., such that the minimum attenuation of the GFF is 0 dB. Connector losses of the GFF stage are also neglected. The GFF case is considered with a flat system input as well as input which is optimized to maximize the minimum SNR on the link.

\begin{figure*}[!t]
\centering
% \begin{mdframed}[hidealllines=true,innerleftmargin=0pt,innerrightmargin=0pt,innertopmargin=0pt,innerbottommargin=0pt,backgroundcolor=yellow]
\subfigure[]{
 \includegraphics[trim=0 0 0 0cm, width=0.27\linewidth]{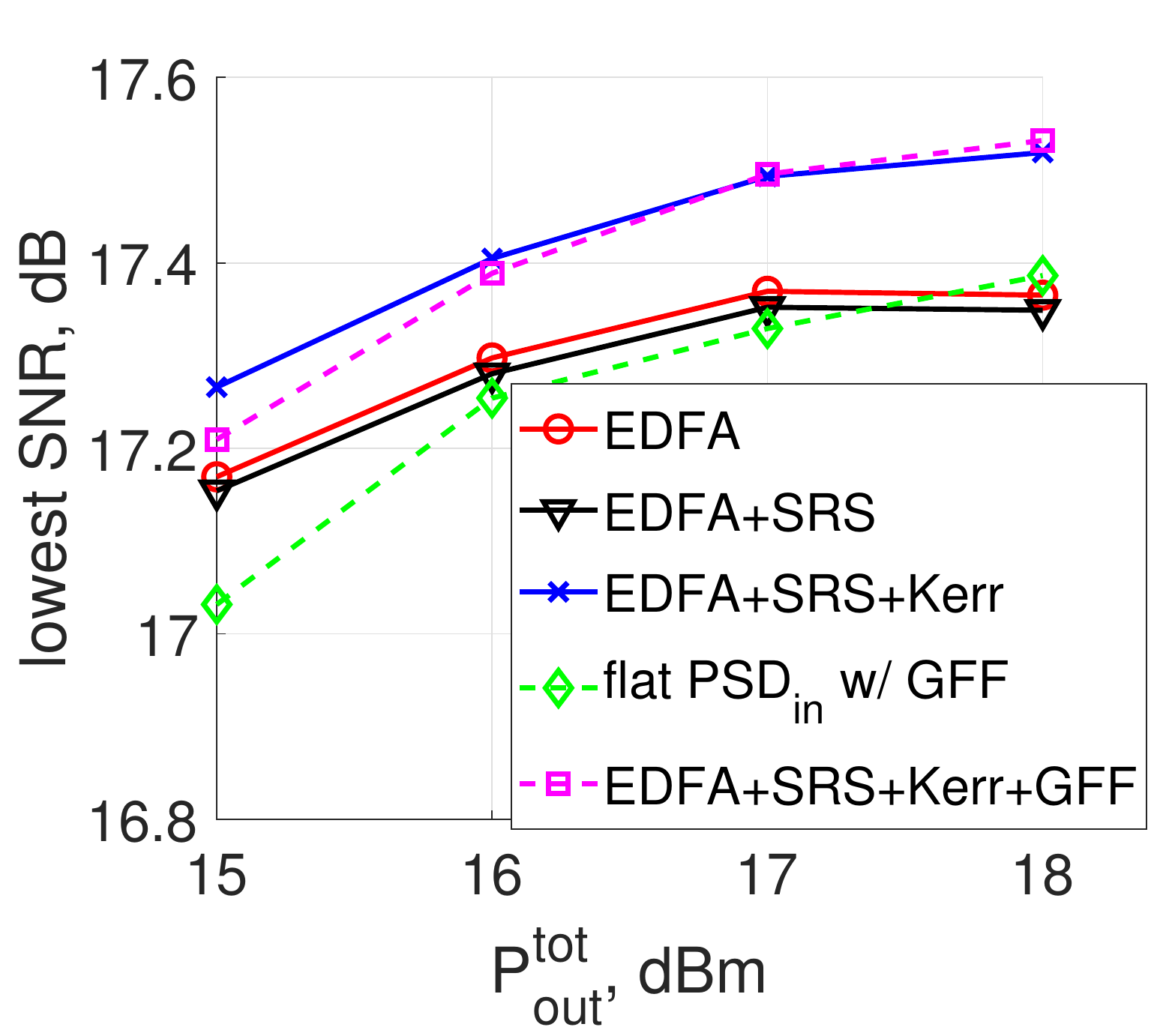}}
\subfigure[]{
 \includegraphics[trim=0 0 0 0cm, width=0.27\linewidth]{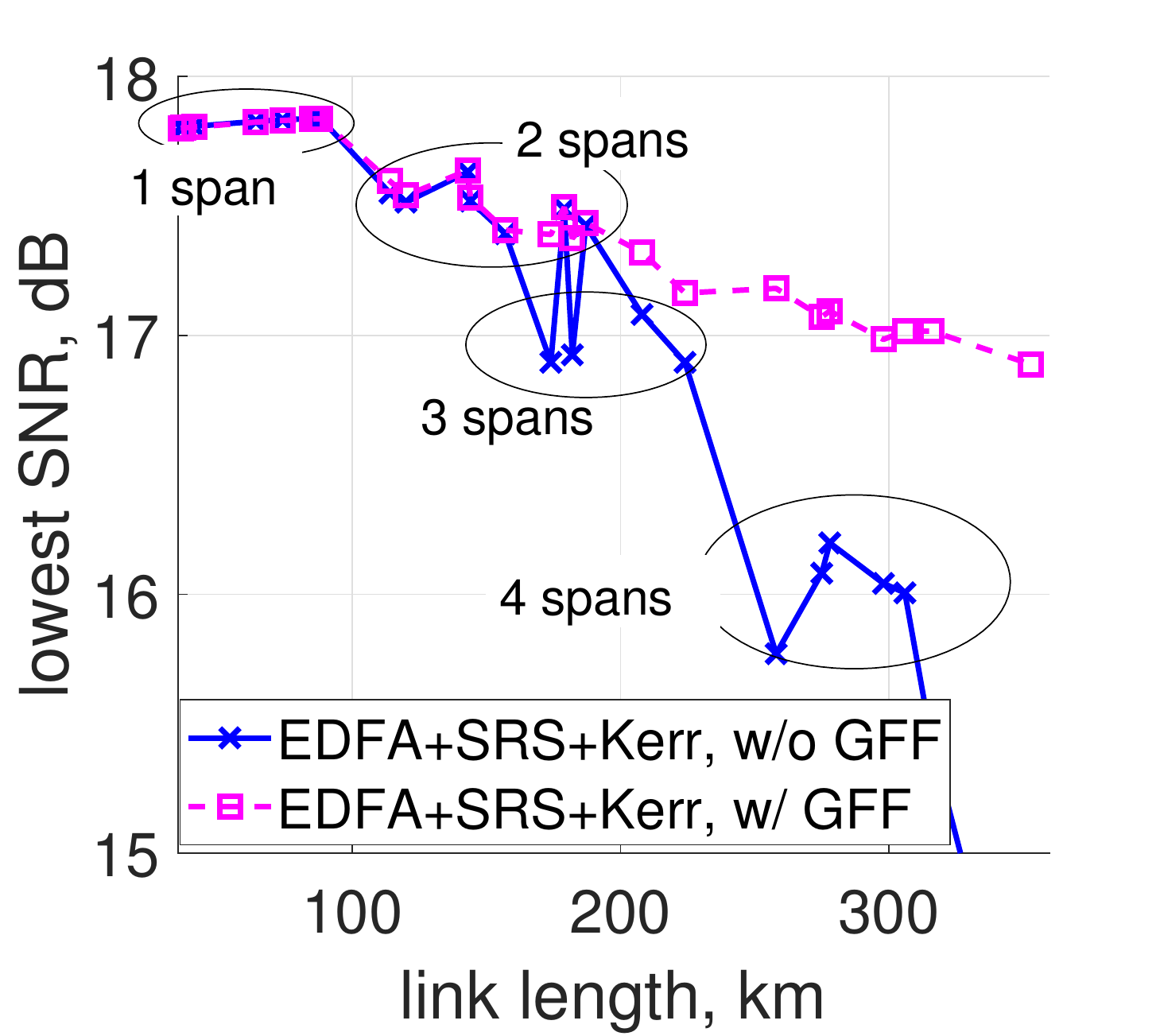}}
\subfigure[]{
 \includegraphics[trim=0 0 0 0cm, width=0.4\linewidth]{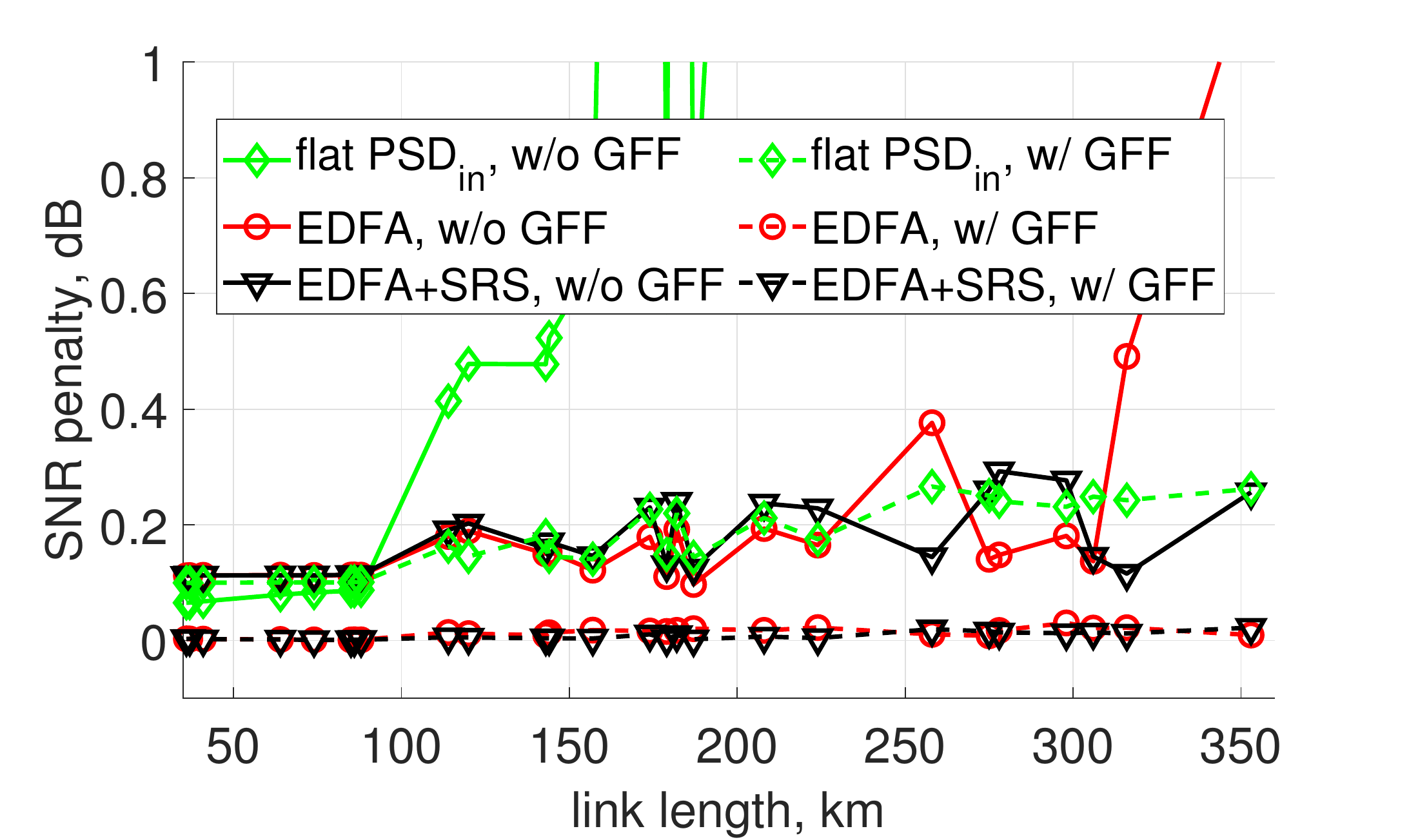}}
 \caption{Performance of the optimizied profiles for the German core network topology.  \textbf{a):} as a function of total launch power for link index 12;  \textbf{b):} comparison between GFF-free and flattened amplifiers with all fiber nonlinearities taken into account;  \textbf{c):} penalty w.r.t. the curves in \textbf{b)} of 1) a flat profile; 2) when optimized only for compensating for the EDFA gain and NF; and 3) when the SRS is included, but not the Kerr nonlinearities.}
 \label{fig:German_network_perf}
%  \end{mdframed}
%  \vspace{-0.4cm}
\end{figure*}

An example of the performance as a function of the total output power is given in \fig{German_network_perf}a) for link index 12 with span lengths 86 and 58 km, respectively. At the maximum output power of the studied amplifier, the performance already approaches its maximum. Even at such short distances, neglecting the Kerr nonlinearities can lead to a performance degradation of $\approx 0.2$ dB. A GFF solution with a flat input spectrum is suboptimal since while it flattens the SRS, it does not capture the Kerr nonlinearities. It can be further improved by optimization and achieve the same performance as a GFF-free solution with an optimized input profile. 
% \end{mdframed}

\begin{table}[!t]
% \begin{mdframed}[hidealllines=true,innerleftmargin=0pt,innerrightmargin=0pt,innertopmargin=0pt,innerbottommargin=0pt,backgroundcolor=yellow]
 \caption{Span length distribution for the links as indexed in Fig.~\ref{fig:GermanTopology}.}
 \label{tbl:spanLenghtDistribution}
 \centering
 \begin{tabular}{c||c|c|||c||c|c}
 \# & Total, km & Spans, km & \# & Total, km & Spans, km\\
 \hline
1 & 36 & {36} & 14 & 174 & {60, 75, 39} \\  
  2 & 37 & {37} & 15 & 179 & {91, 88} \\
  3 & 41 & {41} & 16 & 182 & {68, 80, 34} \\
  4 & 64 & {64} & 17 & 187 & {97, 90} \\
  5 & 74 & {74} & 18 & 208 & {88, 64, 56} \\
  6 & 85 & {85} & 19 & 224 & {85, 95, 44} \\
  7 & 86 & {86} & 20 & 258 & {78, 57, 58, 65} \\
  8 & 88 & {88} & 21 & 275 & {63, 93, 63, 56} \\
  9 & 114 & {73, 41} & 22 & 278 & {84, 83, 59, 52} \\
 10 & 120 & {83, 37} & 23 & 298 & {70, 89, 90, 49} \\
 11 & 143 & {63, 80} & 24 & 306 & {98, 52, 72, 84} \\
 12 & 144 & {86, 58} & 25 & 316 & {56, 75, 98, 87} \\
 13 & 157 & {98, 59} & 26 & 353 & {62, 88, 63, 76, 64} \\
 \hline
 \hline
 \end{tabular}
%  \end{mdframed} 
\end{table}

% \begin{mdframed}[hidealllines=true,innerleftmargin=0pt,innerrightmargin=0pt,innertopmargin=0pt,innerbottommargin=0pt,backgroundcolor=yellow]
The minimum SNR for all links is given in \fig{German_network_perf}b) as a function of the distance at the launch power of $18$ dBm for the case when all fiber nonlinearities are taken into account. For the links in the network up to 3 spans (220 km), the proposed method enables removing the GFF while retaining the same performance at lower power consumption. At the short distances, the unoptimized GFF solution is even penalized ($\approx 0.15 dB$) due to the extra amplification required. However, when the distance is increased, periodic gain flattening becomes necessary. 

The penalty of disregarding the fiber nonlinearities is given in \fig{German_network_perf}c). When a GFF is applied, a flat input is penalized by up to $0.3$ dB. In this case and up to the studied distances, the fiber nonlinearities are not detrimental to the performance, and optimization which compensated for the EDFA is already sufficient. However, without a GFF, disregarding Kerr nonlinearities results in $\approx 0.2$ dB of penalty. For the long distances, disregarding the SRS also has a significant impact on the performance. A GFF-free solution is not applicable without some input power optimization which at least targets the EDFA response, except for single-span systems. 

\section{Discussion and future work}
The network simulations suggest $\approx 0.2$ dB of gain by accounting for the nonlinearities in the optimization of the system, which was not observed in the experiment. We speculate that is due to the experimental uncertainties, which mask the relatively small gain.

It is of general interest to extend the experimental study to EDFAs with GFF, where practical insertion losses and filtering penalties will be present. Furthermore, an analysis of the frequency of gain flattening, or in the general case - gain shaping \cite{JuhnoECOC2020} when fiber nonlinearities are present is also of interest. 

As mentioned, ASE loading can be applied to ensure the EDFAs are fully loaded. However, it would be relevant to also train a model which supports switching off sets of channels. 

The proposed method is backwards compatible and can be used for already deployed links or when changes are made to the link (e.g. an update of an amplifier or other components with different spectral characteristics). For both old and newly deployed links, the proposed method can be used as an initial optimizer, which is then improved when operational data become available by e.g. using a feedback control. 

The symbol rate in this work is limited to 32 GBd by the 64GSa/s AWG. Given the current trends in the industry, analyzing the effect of fiber nonlinearities on the optimization is of interest for higher symbol rates.

Extending this study to multi-band transmission is also of obvious interest as the SRS effect is exacerbated for wider bandwidth. 

% \end{mdframed}

\section{Conclusion}

A machine learning-based model was developed to predict the noise figure (NF) and gain of an Erbium doped fiber amplifier (EDFA) based on its input power spectral density (PSD), total input and total output power. The EDFA model was demonstrated to generalize to multiple devices of the same make. The model was then used to predict the spectral density of the noise produced by the amplifier when part of a multi-span optical fiber communication system of arbitrary configuration in terms of number of spans, span length and launch power per span per channel. Predicting the amplified spontaneous emission (ASE) noise, together with a Gaussian noise model-based estimation of Kerr nonlinearities was then employed for the prediction of the signal to noise ratio (SNR) of such arbitrary, experimental multi-span systems. The system model constructed by cascading the models of the basic components (EDFA, fiber, transceiver penalties) was then used to optimize system performance metrics, such as the SNR w.r.t. the input PSD. When optimized for maximized flat SNR profile, the SNR performance of the worst channel was improved by $6$ dB and $4$ dB w.r.t. an unoptimized, flat input power profile for the two demonstrated system configurations, respectively, achieving a flatness of $\approx1.2$dB. 
By building a collection of models for all EDFA types and makes operational in a given system, the proposed method enables the optimization of reconfigurable, non-homogeneous optical networks without relying on gain-flattening. It was also demonstrated that gain flattening is necessary after 3 spans, in which case the proposed method provides up to 0.3 dB of gain w.r.t. a flat input power profile for up to 360 km.

\section*{Acknowledgment}
This project has received funding from the Danish National Research Foundation (CoE SPOC grant no. 123) and the Villum Foundations (VYI OPTIC-AI grant no. 29344). The authors would like to thank Dr. U.C. de Moura and Dr. Y. An for insightful discussions on measuring the noise figure of EDFAs.

\bibliographystyle{IEEEtran}
\bibliography{references}

\end{document}